\documentclass[11pt,a4paper]{article}
\usepackage{times,latexsym}
\usepackage{url}
\usepackage[T1]{fontenc}

\usepackage[acceptedWithA]{tacl2021v1}

\usepackage{graphicx}
\usepackage{colortbl} 
\usepackage{xcolor}   
\usepackage{multirow}
\usepackage{graphicx}
\usepackage{amssymb}
\usepackage{nicematrix}
\usepackage{booktabs}
\usepackage{transparent}
\usepackage[outline]{contour}

\usepackage{xspace,mfirstuc,tabulary}

\newif\iftaclinstructions
\taclinstructionsfalse 
\iftaclinstructions
\renewcommand{\confidential}{}
\renewcommand{\anonsubtext}{(No author info supplied here, for consistency with
TACL-submission anonymization requirements)}
\newcommand{\instr}
\fi

\iftaclpubformat 

\else

\fi

\newcommand{\datasetheaderU}[2]{\shortstack{\\[0.1ex]#1 \\ {\footnotesize{#2}}}}
\newcommand{\modelcategory}[1]{\rotatebox[origin=c]{90}{\footnotesize{\shortstack{#1}}}}

\definecolor{CustomBlue}{RGB}{25, 100, 210}

\title{ROSA: Addressing text understanding challenges\\in photographs via ROtated SAmpling}

\author{
    Hernán Maina$^{1,2}$,
    Guido Ivetta$^{1,3}$, 
    Mateo Lione Stuto$^1$, \\
    \textbf{Julian Martin Eisenschlos$^{1}$},
    \textbf{Jorge Sánchez$^4$}, \and
    \textbf{Luciana Benotti$^{1,2}$}
    \\
    $^1$FAMAF, Universidad Nacional de Córdoba, $^2$CONICET, Argentina \\
    $^3$Fundación Vía Libre \and $^4$Mercado Libre Inc., Argentina \\
    \small \texttt{\{hernan.maina,guidoivetta,mateo.lione.stuto,julian.eisenschlos\}@mi.unc.edu.ar}\\
    \small \texttt{\nolinkurl{jorge.sanchez@mercadolibre.com,luciana.benotti@unc.edu.ar}}
}

\date{}
\begin{document}
\maketitle

\begin{abstract}
Visually impaired people could benefit from Visual Question Answering (VQA) systems to interpret text in their surroundings.
However, current models often struggle with recognizing text in the photos taken by this population. 
Through in-depth interviews with visually impaired individuals, we identified common framing conventions that frequently result in misaligned text.
Existing VQA benchmarks primarily feature well-oriented text captured by sighted users, under-representing these challenges.
To address this gap, we introduce \textit{ROtated SAmpling} ($ROSA$), a decoding strategy that enhances VQA performance in text-rich images with incorrectly oriented text.
$ROSA$ outperforms $Greedy$ decoding by 11.7 absolute points in the best-performing model.
\end{abstract}

\section{Introduction}
\label{sec:intro}

Extracting text from the surroundings is essential for people with visual impairments. Large multimodal models (LMMs), which possess visual and linguistic processing capabilities, have shown impressive results in tasks such as VQA. They offer great potential but still face real-world challenges that limit their effectiveness in assistive applications for this population.

Although advances in deep learning have improved the detection and recognition of text in arbitrary orientations and shapes~\cite{he2016accuratetextlocalizationnatural, he2018MOSceneText,  yao2016SceneTextDH, bazazian2017improvingtextproposalsscene, Ma2018AOSceneTextD, liao2018RSROSceneTextR, zhang2024ASTextD}, such work typically addresses challenges posed by pictures taken by sighted people in natural scenes---such as the diversity of fonts, sizes, colors, and lighting. 
\citet{VizWiz:nearly-real-time-answers} highlight that nearly a quarter of the questions posed by individuals with visual impairments require reading text present in images. Yet, the specific challenges of reading text in pictures taken by people with visual impairments remain an under-researched area.

\begin{figure}[!t]
    \centering
    \includegraphics[width=\linewidth]{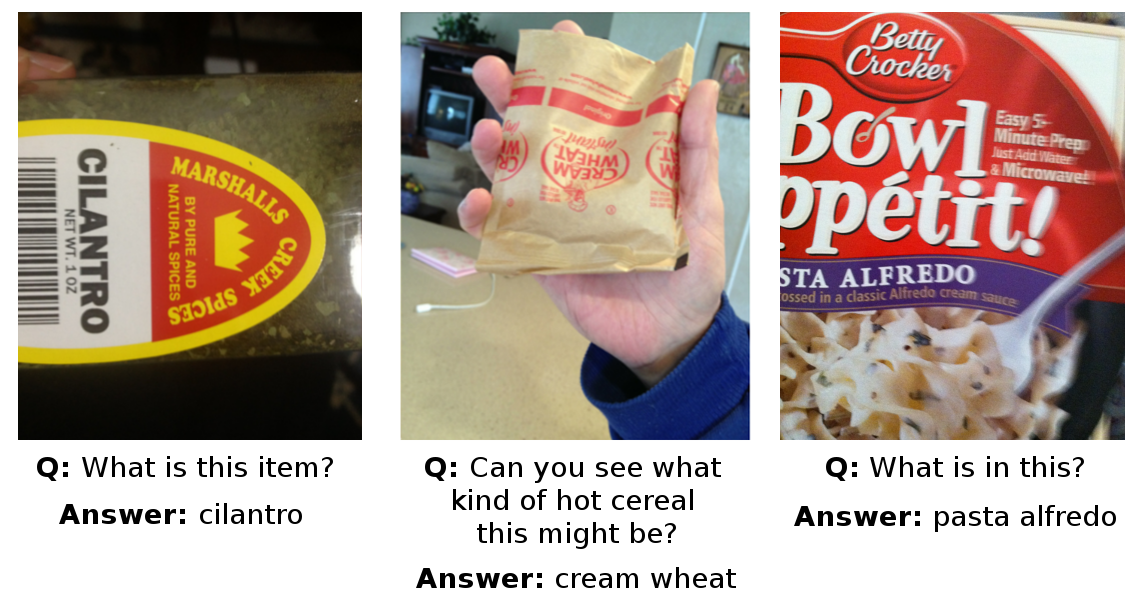}
    \caption{Examples of image pairs and associated visual questions retrieved from the \textit{VizWiz-VQA}~\cite{gurari2018vizwiz} dataset, highlighting challenges such as recognizing text in more than one orientation, addressing occlusions caused by varying surface textures, and handling unconventional framing. These challenges are illustrated from left to right in the three images, respectively.
    }
    \label{fig:vizwiz-txt-samples}
    \vspace{-1em}
\end{figure}

Figure~\ref{fig:vizwiz-txt-samples} illustrates the challenges present in pictures taken by people with visual impairments. The images were extracted from the \textit{VizWiz-VQA}~\cite{gurari2018vizwiz} dataset. The photo on the left illustrate one such challenge: it contains text in different orientations---the yellow text that says \emph{spice} is upside down, while the response is written vertically. As a result, no single rotation can make all the text in the picture correctly oriented.
The image in the middle also contains text which is not correctly oriented with respect to the hand that holds it. As we shall see, people with visual impairments often struggle to correctly orient symmetric objects. This image also presents distorted text and shadows caused by the material of the bag and the way it is being held.
Finally, the image on the right contains only a partial answer---the word \emph{pasta} is only partially visible due to incorrect framing. Sighted people tend to frame pictures by centering the object of interest and leaving some margin around it, but this is not easy to achieve without sight, as our interview process confirms.
 
In this paper we pose three main questions: \textbf{(1)} \textit{How do incorrectly oriented text affect the performance of models in VQA tasks?}, \textbf{(2)} \textit{What strategies can mitigate these limitations?}, and \textbf{(3)} \textit{What factors influence the framing of pictures, and their contained texts, when taken by individuals with visual impairments?}.

Our work undertakes the two first questions from complementary perspectives. First, we show that incorrect text orientations in the text that contains the answers significantly impact all models performance, with an average performance drop of approximately 7.7 absolute points when using the conventional $Greedy$ decoding.
Second, we introduce \textit{ROtated SAmpling} ($ROSA$), a simple yet effective strategy that combines sampling-based decoding with image rotations, outperforming conventional $Greedy$ decoding by 11.7 absolute points on the best-performing model. 
To address the third question we perform in depth interview with people with visual impairments.

We evaluate our proposal using subsets of the \textit{VizWiz-VQA} dataset, which consists of pictures taken by visually impaired people during their daily life. Those pictures, together with a question (as illustrated in Figure~\ref{fig:vizwiz-txt-samples}), were sent through an app to a person that could answer the questions. 
The subsets focus on images containing incorrectly oriented text. The orientations are based on interviews with blind and low-vision individuals, providing insights into how intuitive strategies and conventions learned by them influence the way they frame and capture pictures.

It is important to emphasize that the visual perception of individuals with visual impairments and the rules governing it are shaped through interactions with their environment and experiences provided by family members, educators, and other mediators~\cite{Lewis1969-LEWCAP-4, LUDIKOVA2012971, MAJEROVA2017751}.

Our main contributions are as follows:
\begin{itemize} 
    \setlength\itemsep{0.1em}
    \setlength\topsep{0.1em}
    \item We quantify the weaknesses of state-of-the-art multimodal models in VQA tasks that require answering questions about photographs where reading incorrectly oriented text is necessary.
    \item We identified intuitive patterns and conventions learned by people with visual impairments regarding the orientation of objects and their textual elements, and how these influence the way they frame and capture photographs. Based on these findings, we created evaluation subsets derived from the \textit{VizWiz-VQA} dataset, incorporating systematic variations through rotated images.
    \item We design and evaluate \textit{ROtated SAmpling} ($ROSA$), a decoding strategy that combines sampling-based generation with image rotations. Our results demonstrate its effectiveness in consistently improving the performance of several state-of-the-art multimodal models on images containing incorrectly oriented text.
\end{itemize}

The remainder of the article is structured as follows: Section~\ref{sec:related_work} reviews prior Optical Character Recognition (OCR)-based VQA models, and techniques relevant to our proposal. 
Section~\ref{sec:insights_from_interviews} presents key findings from interviews with visually impaired individuals on their challenges with text in images.
Section~\ref{sec:methods} details the evaluation datasets, informed by the interviews, and the metrics used. 
Section~\ref{sec:rosa_strategy} introduces our $ROSA$ strategy in detail, its operation, and design motivation. 
Section~\ref{sec:analysis_and_results} presents and analyzes the experimental results, including the discussion of the interview insights and the $ROSA$'s performance comparison to other strategies.
Finally, Section~\ref{sec:conclusions} concludes the paper, highlighting the importance of visually impaired individuals' needs in VQA design.
\section{Related work}
\label{sec:related_work}

VQA models designed to read and understand text in images have received increasing attention in recent years. Early approaches, such as \textit{LoRRA}~\cite{singh2019towards}, extended standard VQA architectures by incorporating Optical Character Recognition (OCR) tokens into the answer generation process. Subsequent works, like \textit{M4C}~\cite{hu2020iterative}, introduced multimodal fusion mechanisms to better integrate text and image features, improving performance on datasets such as \textit{TextVQA}~\cite{singh2019towards}, \textit{ST-VQA}~\cite{ST-VQA:dataset} and \textit{OCR-VQA}~\cite{mishraICDAR19}. More recently, models such as \textit{TAP}~\cite{Yang2021TAP} and \textit{Pix2Struct}~\cite{lee2023pix2struct} have leveraged transformer-based architectures and pretraining on large-scale datasets to enhance OCR-based VQA capabilities. However, these models are typically trained and evaluated on curated datasets where text is well-aligned and clearly legible, limiting their robustness to real-world challenges such as those found in images captured by visually impaired individuals.

Regarding VQA model training and inference, we distinguish between \textit{data augmentation in training} and \textit{inference-time ensembles}. We describe related work for each of them below and we situate this paper contribution regarding them.  

\paragraph{Data augmentation in training} is a widely adopted technique in Computer Vision (CV) to improve the robustness and generalization of models~\cite{shorten2019survey}.
By artificially expanding the diversity of the training dataset with label-preserving transformations, these methods aim to prevent overfitting and enhance model performance.
Common data augmentation techniques include geometric transformations such as rotations, translations, scaling~\cite{Simonyan15}, cropping, and flipping, which help models to become invariant to spatial variations in the input images. In addition to geometric transformations, \citet{krizhevsky-et-al-2012-imagenet} also employed random adjustment of color balance, demonstrating significant improvements in image classification tasks.
\citet{engstrom2019a} showed how many image classification models were not robust to rotations in the input images, further demonstrating the need for these methods.

Although these techniques focus on training-time methods, they set the foundation for exploring augmentation strategies beyond the training phase, such as inference-time augmentation. 

\paragraph{Inference-time ensembles} leverage (or boost) the variability in predictions to improve reliability and performance. 
Techniques such as Monte Carlo Dropout~\cite{pmlr-v48-gal16} simulate an ensemble by performing stochastic forward passes with dropout activated during inference. This approach captures uncertainty by sampling over a distribution of possible models in a Bayesian fashion. 
Another recent strategy involves adding variability through self-consistency prompting~\cite{wang2023selfconsistency}, where multiple predictions are generated by introducing controlled randomness into the sampling process. By aggregating these predictions, the ensemble mitigates errors from individual predictions.

Our approach $ROSA$ (see details in Section~\ref{sec:rosa_strategy}) combines the ideas from image augmentation to build better inference-time ensembles. Sampling from image augmentations provides an efficient way to sample from a single black-box model and simultaneously tackle a weakness in the performance of vision models on images taken by people with visual impairments.
\section{Insights from in-depth interviews}
\label{sec:insights_from_interviews}

Our in-depth interviews with visually impaired people highlight their challenges in reading text from natural images due to the lack of visual perception and limited interaction with image-capturing technologies. In this section, we introduce the participants and provide quotes to contextualize our findings.

We interviewed eight participants (two women, six men), aged between 23 and 70 years, with a mean age of 38. Four participants with total blindness and four with partial vision. All had completed secondary education, and five held higher education degrees. Their professional backgrounds included fields such as law, psychology, and the arts. Participants were recruited through special education institutions and independent outreach efforts, ensuring a diverse sample with varying degrees of visual impairment.

To contextualize the interview, participants were asked to answer a series of questions about their technology use and their demographics via an online form designed for full accessibility. 
We asked \textit{How often and for what purposes do you use your phone’s camera?}. Participants reported a wide range of camera usage, from occasional to daily, serving both practical and personal needs. Some use it sparingly for documents, family moments, or landscapes, while others rely on it more frequently to read fine print, check labels, or magnify details. Additional uses include text scanning, verifying banknotes, and taking pictures for others. Among those with total blindness, only one used the camera frequently, while others faced significant challenges in setting up photographs, with one never having taken a picture before.

One participant noted that familiarity with one's environment plays a key role in navigating visual challenges:
\textit{``People with blindness, whether from birth or acquired, become very dependent on the environment in which they were raised''}. This demonstrates how spatial memory and learned visual conventions aid daily activities, though temporary objects like food packaging and hygiene items remain difficult to identify.

One of the principal barriers identified during the interviews was the lack of a universal standard for identifying small objects, such as on product labels or packaging, which complicates accessibility to visual information, highlighting the need to develop solutions tailored to these diverse conventions and experiences. The lack of standardized tactile markers increases accessibility challenges for individuals with visual impairments.
One participant explained while referring to the task of identifying an object: \textit{``if the element does not have aids or conventional patterns to identify them, the task becomes even more difficult, a task that is already quite complex''}. 

Additionally, participants noted that even when technology provides assistance, the way information is conveyed can still pose barriers. Specifically, using relative referring expressions such as ``the label is at the bottom'' or ``the button is on the left'' can be ineffective for blind users. Unlike sighted individuals, who can quickly orient themselves based on visual landmarks, people with visual impairments rely on tactile exploration and fixed object features. Without a clear reference point, terms like bottom or left become ambiguous. Since they frequently do not know which way the object is supposed to be oriented, telling them something like \emph{``at the bottom''} does not help if they are holding it upside down. This highlights the need for more precise descriptions, such as referencing distinct textures, shapes, or absolute positioning (e.g., \emph{``near the corner with raised dot'')}. 

Papers, packages, and containers often include text that identifies them---such as those in Figure~\ref{fig:vizwiz-txt-samples}--- but, due to the lack of universal patterns and the absence of clear tactile features to identify visual structures, visually impaired people cannot read it without technological aids. Many essential objects lack Braille labels, making independent identification difficult. This is especially problematic for medication packaging and supermarket products---one of the interviewees describes this difficulty by saying \textit{``It is a huge challenge trying to identify products that come sealed in security packages because I can’t touch them.''}. 

One finding important for this work is that objects that they need to identify have different symmetries that can be detected through touch, suggesting that some rotations of the objects are indistinguishable for people with visual impairments. This paper focuses on examining small, movable objects where text is crucial for usability and identification.
\section{Methods}
\label{sec:methods}

This section outlines the design, processing, and evaluation of the datasets used in this work's experiments. Building on the \textit{VizWiz-VQA} dataset and adopting a community-based approach, we incorporated conventions observed during interviews with visually impaired people to create datasets specifically addressing the challenge of incorrectly oriented text. Additionally, we describe the evaluation metrics aplplied.

\subsection{Dataset for the visually impaired}
\label{sub-sec:dataset-for-the-visually-impaired}

The base dataset used in this research is \textit{VizWiz-VQA}, a resource derived from the \textit{VizWiz}~\footnote{https://vizwiz.org/} project, which focuses on creating datasets and challenges aimed at developing assistive technologies for people with visual impairments.
\textit{VizWiz-VQA} consists of question-image pairs generated by visually impaired people during their daily activities, each of which is associated with 10 answers provided by crowd workers.
Unlike traditional VQA datasets, it includes images with challenges like blurriness, low resolution, and other technical defects, which pose significant obstacles as illustrated in~Figure \ref{fig:vizwiz-txt-samples}.

To assess the performance of the models in text reading and recognition tasks within images, only \textit{VizWiz-VQA} cases where these skills were crucial for answering the questions were selected.\footnote{Only the original \textit{training} and \textit{validation} datasets were used, as the \textit{test} set is not publicly available.}

The selection process consisted of two phases. In the first phase, pairs were discarded if ``unanswerable'' was the most frequent response due to unclear questions, insufficient visual evidence, or poor image quality. In the second phase, pairs were filtered based on annotations from the work of \citet{ZengWCBG20}, keeping only those in which at least three of the five involved annotators agreed that text reading and recognition were the required skills to correctly answer the questions.

The final dataset consisted of approximately 6300 samples, from which 1270 (20\%) were randomly selected for the experiments in this work, referred to from here on as \textit{VizWiz-Text}.
Since nearly 20\% of the images in this subset contained incorrectly oriented text, this highlighted the need to investigate how visually impaired people capture images and how to improve dataset representativeness regarding misoriented text. To address this, interviews with individuals with varying visual impairments informed the design of derived datasets for evaluation (see Subsection~\ref{sub-sec:community-informed-augmentation}).

\subsubsection{Derived datsets}
The interviews revealed learned conventions and patterns used by people with visual impairments when orienting objects to be photographed. These findings were crucial for designing the following datasets, which enhance the representativeness of images with incorrectly oriented text.

\paragraph{VizWiz\,$(Conventional)$} represents everyday scenarios where visually impaired individuals may capture images with misoriented text. The dataset includes 990 samples from \textit{VizWiz-Text}, comprising: (1) all pairs with images containing text rotated $\geq|90|^\circ$, and (2) specific rotated variants (at 90°, 180°, and/or 270°) of images with originally well-oriented text. The rotations were selected based on patterns identified through interviews with visually impaired individuals. 

\paragraph{VizWiz\,$(Oriented)$} is a subset of 945 samples from \textit{VizWiz-Text}, where the text in the images is correctly oriented, i.e., rotated $< |90|^\circ$.

\paragraph{VizWiz\,$(Random)$} includes all pairs from \textit{VizWiz\,($Oriented$)}, but with images randomly rotated to 90º, 180º, or 270º. Like \textit{VizWiz\,($Conventional$)}, it simulates misoriented text but focuses on technical issues, such as disabled auto-rotation, rather than challenges specific to visual impairment. This introduces greater variability in image orientations.

\subsubsection{Baseline datasets}
As a reference dataset, we used the \textit{OCR-VQA}~\cite{mishraICDAR19}, which contains approximately 1.03 million pairs of book cover images and associated questions. Unlike \textit{VizWiz-Text}, this dataset not present issues related to image legibility or quality, and 100\% of its images contain correctly oriented text, providing ideal conditions for evaluating model performance. For our experiments, we used a subset of 1270 samples randomly selected from the \textit{test} partition, which we will refer to as \textbf{OCR-VQA\,$(Oriented)$}. Additionally, following a similar approach to that used for \textit{VizWiz\,$(Random)$}, we generated a rotated version of this dataset, named \textbf{OCR-VQA\,$(Random)$}.

\subsection{Community-informed augmentation}
\label{sub-sec:community-informed-augmentation}

The interviews explored how participants perceive and orient everyday objects with printed text. To simplify execution, the task was adapted to a 2D plane, with objects placed on a table in predetermined positions. Each interview involved 14 objects: 12 for the main task and 2 for practice. Objects were selected for diverse characteristics, including tactile guides, symmetry, and complexity in text orientation. The interviewer placed each object in four positions (0º, 90º, 180º, and 270º), ensuring the written response was visible. The interviewee, seated next to the interviewer, was informed that only one position displayed the text in the correct orientation ---that is, left to right and unrotated.

Participants assessed text orientation by answering ``Yes'' (text is readable), ``No'' (text is misoriented), or ``I cannot know'' (uncertain due to object characteristics), with a ``Yes'' response ruling out ``I cannot know''. For example, if the object was a cereal box, the question might be: \textit{“What is the brand of this cereal box?”}. Participants had to determine whether the text was oriented correctly for the interviewer to read. To standardize vision levels and focus on learned conventions, all participants wore blindfolds. The practice phase included immediate feedback to clarify instructions, while the main phase prioritized analyzing perception. Participants were also encouraged to explain their reasoning, providing deeper insight into their strategies.

\subsection{Evaluation metrics}
\label{sub-sec:evaluation-metrics}

To assess the performance of the models on the VQA task, we used three main metrics: \textit{Exact Match} (EM), \textit{Standard VQA Accuracy} (Acc), and \textit{Semantic Answer Similarity} (SAS). Before computation, both model predictions and reference answers underwent standardized preprocessing, which included lowercase conversion, article removal, contractions correction, punctuation replacement with spaces (except apostrophes and colons), and trimming of spaces.

The \textit{EM} metric follows the standard exact matching approach, while \textit{Acc} is computed as defined in \citet{Antol_2015_ICCV}, where a generated answer is considered correct if it matches at least three reference answers. The \textit{SAS} metric evaluates semantic similarity using the \textit{all-MiniLM-L6-v2} model~\cite{wang2020minilm}, which generates vector embeddings to capture semantic content. Similarity is measured via cosine similarity between vectors, offering a more flexible and human-aligned evaluation.
\section{ROtated SAmpling}
\label{sec:rosa_strategy}

The \textit{ROtated SAmpling} ($ROSA$) strategy aims to improve the ability of multimodal models to interpret text in images with variable orientations. The core idea is to present the model with multiple rotated versions of the same image, each associated with a given question. These rotations systematically alter the orientation of the text within the image, allowing the model to process the input from different perspectives. For each rotation, the model generates multiple independent predictions, each assigned a confidence score that reflects the probability of correctness.

\paragraph{Inference process.} 
For each inference, we use the following prompt format: ``\texttt{Answer the question using a single word or phrase. <question>}''.
This format was inspired by the prompt style used by \citet{Liu_2024_CVPR} for evaluating the \textit{VizWiz-VQA} dataset. It helps ensure concise and relevant answers in the expected format. However, models fine-tuned on the tested datasets did not require this prompt, as they had already been trained to interpret questions and generate appropriate responses.

We adopt a \textit{sampling decoding} strategy, setting the \textit{temperature} parameter to $0.5$, while \textit{top\_p} and \textit{top\_k} are set to $1.0$ and $50$, respectively. This configuration strikes a balance between variability and coherence, allowing the model to explore a broader range of plausible answers while maintaining response consistency.

The final prediction returned by $ROSA$ corresponds to the answer (i.e., inference) with the highest confidence likelihood-based score. Let $v$ and $q$ be the associated image and question, respectively. The likelihood of a given answer $a$, denoted as $p(a|v,q)$, is computed using the chain rule as follows: $p(a|v,q) = \prod_{i=1}^n p(t_i | t_1, \cdots, t_{i-1}, v, q)$, where $t_i$ are the tokens that form the answer $a$. This formulation represents the probability assigned by the model to generating the sequence of tokens in $a$, given both the visual and textual inputs. In case of ties among the top $k$ predictions, we apply majority voting to select the most frequent answer. If a tie remains after voting, the final answer is randomly selected from the set of tied predictions.

Unlike the traditional $Greedy$ decoding approach, which selects the most likely answer from a single inference about the original image, our strategy leverages multiple perspectives, which improves robustness when working with incorrectly oriented text.

\paragraph{Motivation and rationale.}
The design of $ROSA$ is grounded in the characteristics of the datasets commonly used to train multimodal models that process both visual and textual inputs. Many of these models are trained on artificial datasets where, although rotated text may be present, standard orientations---such as left-to-right horizontal text---predominate. Given this bias in the training data, it is reasonable to assume that models tend to develop a stronger affinity for such configurations. Consequently, it is expected that predictions made on images with correctly oriented text will, on average, be associated with higher confidence scores.

\begin{figure}[!t]
    \centering
    \includegraphics[width=\linewidth]{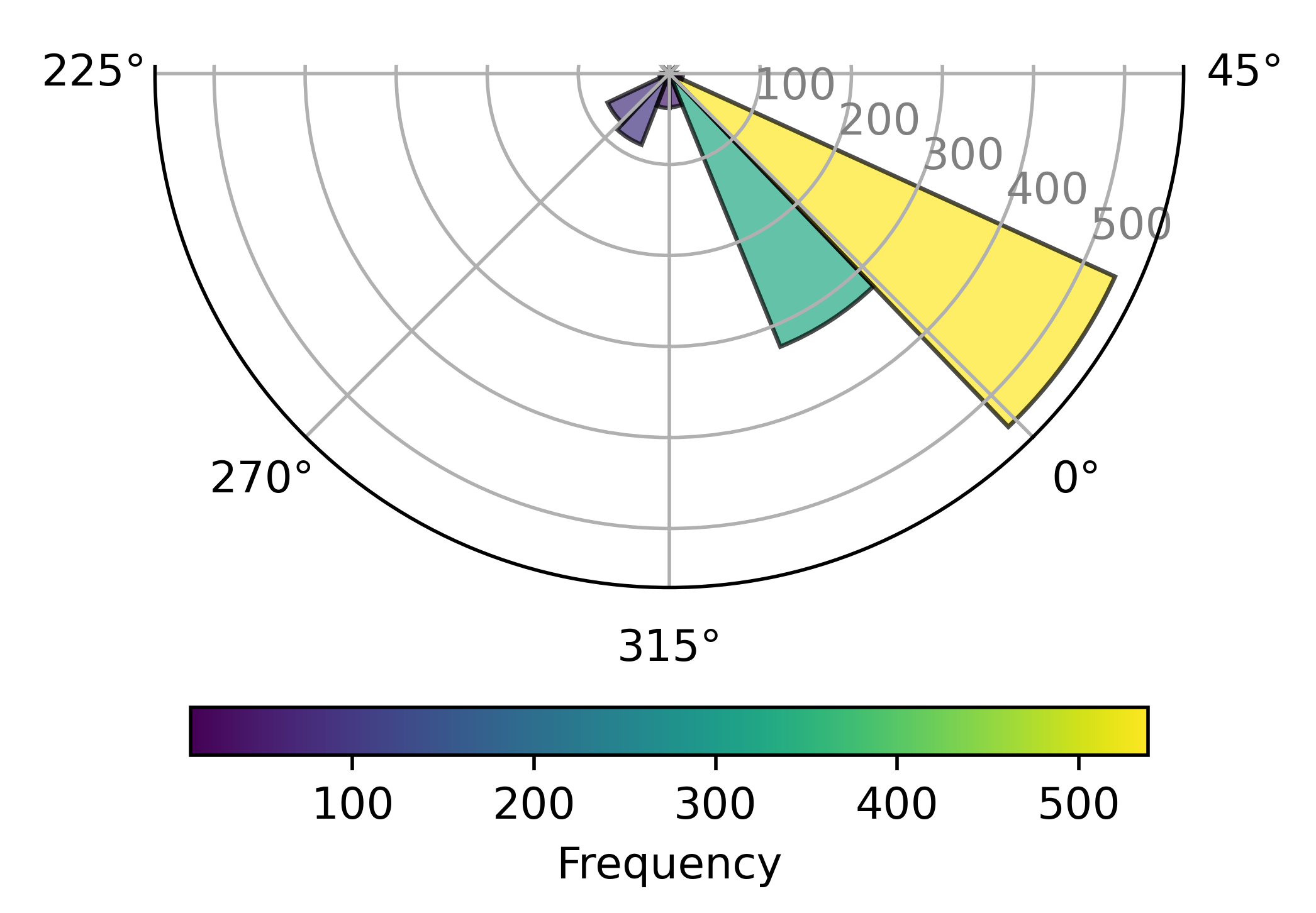}
    \caption{
    Polar histogram of text orientations in the \textit{VizWiz-Text} dataset. The most common orientations were 0° (landscape) and 270° (portrait), while no text was detected at 90° or 180° due to the model's limitations in recognizing symmetric orientations.
    }
    \label{fig:vizwiz_test_polarplot}
    \vspace{-1em}
\end{figure}

To implement this approach, we selected four key rotations: 0°, 90°, 180°, and 270°. This decision was based on an empirical analysis conducted on the 1270 images that make up the \textit{VizWiz-Text} dataset. We employed a state-of-the-art text detection model~\cite{zhang2024ASTextD} and computed the main orientation axis of the detected text in the image.

The analysis (Figure~\ref{fig:vizwiz_test_polarplot}), showed that most texts were aligned in landscape (0°) or portrait (270°) orientations. However, due to limitations of the model at the time of testing, it could not distinguish symmetric rotations (vertical and horizontal), leading to the apparent absence of text at 90° and 180°.
Despite this, we assume such orientations exist and chose the four canonical rotations to ensure broad coverage while maintaining computational efficiency.
\section{Analysis and results}
\label{sec:analysis_and_results}

In this section, we start by describing what we learned from the interviews with the participants. We then compare the performance of the models using the standard $Greedy$ strategy, our proposed $ROSA$, and two derived strategies, presenting empirical results on both the interview-informed and reference datasets. Finally, we analyze the impact of text rotations on the performance of the models, with a detailed focus on $ROSA$.

\subsection{Challenges in orienting objects}
Of the approximately 384 questions asked during the interviews, 67\% of the responses aligned with the hypothesized conventions regarding the expected rotations from the perspective of a person without visual impairments. A key finding was the difference in responses based on the degree of visual impairment. Participants with partial vision or prior visual experience tended to follow more conventional patterns similar to sighted people, while those with total blindness from birth presented more diverse responses. We explain insights that we obtained from the errors of the 33\%. In our hypotheses, we classify the objects used in the interview in three groups according to their symmetries or distinguishing tactile features, as we introduced in Section~\ref{sec:methods}. 
Based on the results from the interviews, we first describe the most difficult objects to orientate correctly, followed by those of intermediate difficulty, and finally those easily oriented by visually impaired people.
Below, we outline the insights: the first three align with the categories predicted by the hypotheses and confirmed by the interviews, while the last three are emergent findings.

\paragraph{Multiple symmetries are more difficult.} The symmetry and lack of clear reference points in objects such as circular \textit{coffee capsules}, and square \textit{salt} or \textit{juice packets}, significantly hindered the identification of the correct text orientation, generating uncertainty among the participants.

\paragraph{Tactile cues often don't break all symmetries.} For objects with unique distinctive features that break the symmetry, such as \textit{clothing tags}---that have a hole or attached string---, \textit{toothpaste tubes}---which have a cap---, and \textit{cereal bars}---which have seals---the results in determining orientation were more predictable in most cases. In these cases, participants relied on their prior experiences and distinctive features of the object, such as textures, shapes, or functional elements (e.g., caps, holes, or seals), to reduce the possible orientations for the text to be correctly oriented. But they could not be certain about a single orientation, more than one orientation was possible from their perspective. 

\paragraph{Non-symmetric objects are easier.} Relevant aspect observed was the tendency to associate certain objects with specific positions, based on how these objects were displayed in stores or supermarkets, as evidenced in the case of \textit{battery blister packs}. Additionally, for very everyday objects with very distinctive characteristics, such as \textit{CD boxes}, \textit{books} or \textit{soft drink cans}, identifying the correct orientation was certain for most participants, regardless of their degree of visual impairment.

\paragraph{The aspect ratio plays a role.} The aspect ratio plays a role when the subjects decided on the potential orientations. The predominance of landscape orientation was significant when the length and the width differed significantly. For products such as \textit{business cards} or \textit{pens}, most participants assumed that the text would be horizontally oriented from left to right, based on their general knowledge of space and layout. For individuals with total blindness and knowledge of the Braille writing and reading system, this association was closely linked to what it meant for them for the text to be correctly oriented.

\paragraph{Reliance on useful and hidden tactile cues.} During the interviews, we observed how small details on objects ---such as a wider top seal on a \textit{juice packet} or distinctive upper and lower closure edges on \textit{food boxes} like those for \textit{rice} or \textit{cereal}--- were used by many participants to narrow down possible text orientations, and even to confidently determine the correct one. These findings, which were not accounted for in our initial hypotheses, help partially explain the observed error rates.

\paragraph{Reliance on meaningless tactile cues.} Although tactile cues helped orient many objects, in some cases they led to excessive confidence in incorrect conventions based on the participants' daily activities. For example, in a cylindrical \textit{canned food} container without an easy-open system, one participant assumed that a higher edge on a lid indicated the base, while another relied on the horizontal text convention---both were wrong.
Similarly, \textit{medication boxes} were oriented according to perforated barcode markings associated with storage habits, and double-sealed \textit{food bags}, whose orientation depended on the arrangement of the seals (ranging from horizontal to vertical), led to incorrect orientations that, as before, contributed to the reported error rates.

\subsection{Evaluation of model performances}
\begin{table*}[t]
    \centering
    \footnotesize
    \begingroup
        \resizebox{1.0\textwidth}{!}{%
        \begin{NiceTabular}{@{}c |l|l|l |ccc|ccc|cc||ccc|cc@{}}
    & 
    \multirow{5}{*}{\textbf{Model}} &
    \multirow{5}{*}{\textbf{Size}} &
    \multirow{5}{*}{\shortstack{\textbf{Decoding} \\ \textbf{Strategy}}} &
    \multicolumn{8}{c|}{\textit{Images with \textbf{incorrectly} oriented text}} & 
    \multicolumn{5}{c}{\textit{Images with \textbf{correctly} oriented text}} \\[-0.5ex]
    \cline{5-17}

    & & & &
    \multicolumn{3}{c}{\datasetheaderU{\textbf{VizWiz}}{$(Conventional)$}} &
    \multicolumn{3}{c}{\datasetheaderU{\textbf{VizWiz}}{$(Random)$}} &
    \multicolumn{2}{c|}{\datasetheaderU{\textbf{OCR-VQA}}{$(Random)$}} &
    \multicolumn{3}{c}{\datasetheaderU{\textbf{VizWiz}}{$(Oriented)$}} &    
    \multicolumn{2}{c}{\datasetheaderU{\textbf{OCR-VQA}}{$(Oriented)$}} \\[0.5ex]
    \cline{5-17}

    & & & & \scriptsize{\textit{EM}} & \scriptsize{\textit{Acc}} & \scriptsize{\textit{SAS}} &
        \scriptsize{\textit{EM}} & \scriptsize{\textit{Acc}} & \scriptsize{\textit{SAS}} &
        \scriptsize{\textit{EM}} & \scriptsize{\textit{SAS}} &
        \scriptsize{\textit{EM}} & \scriptsize{\textit{Acc}} & \scriptsize{\textit{SAS}} &
        \scriptsize{\textit{EM}} & \scriptsize{\textit{SAS}} \\
    
    \hline \\[-1ex]

    \multirow{8}{*}{\modelcategory{Open weights + data}} &
        \textit{LLaVA-1.6-Vicuna-7B$_{(4bit)}$} & 7B &
            $Greedy$ & 
                17.3 & 26.5 & 53.3 & 
                21.2 & 28.8 & 55.3 & 
                48.3 & 73.6 & 
                {34.7} & {45.3} & {67.6} & 
                66.7 & {88.5} \\

            &~\cite{liu2024llavanext}&&
            $RepeatRot$ & 
                 17.6 &  26.9 &  53.6 &
                 21.5 &  29.5 &  55.3 &
                 48.6 &  73.6 &
                 \textbf{34.8} &  \textbf{45.6} &  \textbf{67.8} & 
                 \textbf{67.0} &  \textbf{88.6}\\

            &&&$RandomRot$&
                 22.7 &  33.4 &  56.9&
                 26.3 &  35.1 &  58.9 &
                 58.8 &  82.9 &
                 31.4 &  41.7 &  62.9 & 
                 66.8 &  87.8 \\
            
            &&&\cellcolor{magenta!5}\textcolor{magenta}{\textit{\textbf{ROSA}}} & 
                \cellcolor{magenta!5}\textbf{27.6} & \cellcolor{magenta!5}\textbf{39.8} & \cellcolor{magenta!5}\textbf{61.3} &
                \cellcolor{magenta!5}\textbf{31.9} & \cellcolor{magenta!5}\textbf{41.6} & \cellcolor{magenta!5}\textbf{63.0} & 
                \cellcolor{magenta!5}\textbf{65.3} & \cellcolor{magenta!5}\textbf{87.3} & 
                \cellcolor{magenta!5}32.0 & \cellcolor{magenta!5}41.8 & \cellcolor{magenta!5}63.1 & 
                \cellcolor{magenta!5}{65.6}$^{\dagger}$ & \cellcolor{magenta!5}{87.5}$^{\dagger}$ \\

        \\[-1ex] & \textit{Molmo-7B-D$_{(bf16)}$} & 7B &
            $Greedy$ &
                24.7 & 33.9 & 61.2 &
                26.8 & 35.2 & 61.9 &
                31.5 & 61.3 &
                {39.0} & {51.0} & {73.8} &
                42.0 & 71.8 \\

            &~\cite{deitke2024molmopixmoopenweights}&&
            $RepeatRot$ & 
                 25.8 &  34.9 &  61.9 &
                  27.6 &  36.3 &  62.6 &
                 36.3  &   66.6 &
                 \textbf{39.5} &  \textbf{51.5} &  \textbf{74.2} & 
                 \textbf{47.4}  &  \textbf{77.2} \\

            &&&$RandomRot$&
                 32.2 &  44.1 &  68.2 &
                 33.9  &  44.5 &  67.7 &
                 39.9 &  72.4  &
                 35.1 &  45.7 &  68.5&
                 {43.2}&   74.4\\
                
            &&&\cellcolor{magenta!5}\textcolor{magenta}{\textit{\textbf{ROSA}}} & 
                \cellcolor{magenta!5}\textbf{36.0} & \cellcolor{magenta!5}\textbf{48.6} & \cellcolor{magenta!5}\textbf{70.6} &
                \cellcolor{magenta!5}\textbf{37.8} & \cellcolor{magenta!5}\textbf{48.9} & \cellcolor{magenta!5}\textbf{70.7} &
                \cellcolor{magenta!5}\textbf{41.7} & \cellcolor{magenta!5}\textbf{74.5} &
                \cellcolor{magenta!5}37.8 & \cellcolor{magenta!5}48.9 & \cellcolor{magenta!5}70.6 &
                \cellcolor{magenta!5}41.8 & \cellcolor{magenta!5}74.5 \\

    \hline \\[-1ex]

    \multirow{16}{*}{\modelcategory{Open weights only}} &
        \textit{Llama-3.2-Vision-11B-Instruct$_{(4bit)}$} & 11B &
            $Greedy$ &
                30.6$^{*}$ & 41.4$^{*}$ & 69.5$^{*}$ &
                25.2 & 34.1 & 61.4 &
                29.8 & 59.7 &
                31.0$^{*}$ & 41.0$^{*}$ & {67.5}$^{*}$ & 
                31.6 & 61.5  \\

            &~\cite{grattafiori2024llama3herdmodels} &&
            $RepeatRot$ & 
                 31.6 &  42.6 &  70.3 &
                 26.5 &  36.0 &  62.4 &
                 32.8  &  62.3 &
                 \textbf{32.7} &  43.2 &  \textbf{68.4} & 
                 34.6 &  64.4 \\
                
            &&&$RandomRot$ & 
                 35.2$^{*}$ &  46.7$^{*}$ &  72.6$^{*}$ &
                 30.9$^{*}$ &  40.8$^{*}$ &  65.8$^{*}$ &
                 35.0 &  64.0 &
                 {32.6}$^{*}$ &  \textbf{43.7}$^{*}$ &  66.7$^{*}$ &
                 36.7 &  64.9 \\
                
            &&&\cellcolor{magenta!5}\textcolor{magenta}{\textit{\textbf{ROSA}}} &
                \cellcolor{magenta!5}\textbf{35.9}$^{*}$ & \cellcolor{magenta!5}\textbf{46.8}$^{*}$ & \cellcolor{magenta!5}\textbf{73.7}$^{*}$ &
                \cellcolor{magenta!5}\textbf{31.1}$^{*}$ & \cellcolor{magenta!5}\textbf{42.3}$^{*}$ & \cellcolor{magenta!5}\textbf{66.7}$^{*}$ &
                \cellcolor{magenta!5}\textbf{36.9} & \cellcolor{magenta!5}\textbf{65.5} &
                \cellcolor{magenta!5}30.8$^{*}$ & \cellcolor{magenta!5}41.8$^{*}$ & \cellcolor{magenta!5}66.5$^{*}$ &
                \cellcolor{magenta!5}\textbf{37.0} & \cellcolor{magenta!5}\textbf{65.6} \\

        \\[-1ex] & \textit{Qwen2.5-VL-3B-Instruct} & 3B &
            $Greedy$ & 
                45.1$^{*}$ & 56.1$^{*}$ & 76.5$^{*}$ &
                46.0 & 55.5 & 74.8 & 
                67.9 & 87.9 &
                {56.4}$^{*}$ & {67.4}$^{*}$ & {82.6}$^{*}$ &
                74.9 & {92.0} \\

            &~\cite{qwen2.5-VL}&&
            $RepeatRot$ & 
                 46.2$^{*}$ &  57.8$^{*}$ &  77.3$^{*}$ &
                 47.1 &  56.8 &  75.2 &
                 69.1 &  88.6 &
                 \textbf{57.6}$^{*}$ &  \textbf{68.5}$^{*}$ &  \textbf{82.8}$^{*}$ &
                 \textbf{75.6} &  \textbf{92.2} \\

            &&&$RandomRot$ & 
                 48.3$^{*}$ &  59.7$^{*}$ &  78.0$^{*}$ &
                 46.8$^{*}$ &  58.0$^{*}$ &  75.0$^{*}$ &
                 72.0 &  89.2 &
                 53.7$^{*}$ &  64.3$^{*}$ &  78.9$^{*}$ &
                 75.0 &  91.6 \\

            &&&\cellcolor{magenta!5}\textcolor{magenta}{\textit{\textbf{ROSA}}} & 
                \cellcolor{magenta!5}\textbf{56.8}$^{*}$ & \cellcolor{magenta!5}\textbf{67.7}$^{*}$ & \cellcolor{magenta!5}\textbf{82.6}$^{*}$ &
                \cellcolor{magenta!5}\textbf{55.2}$^{*}$ & \cellcolor{magenta!5}\textbf{65.5}$^{*}$ & \cellcolor{magenta!5}\textbf{79.9}$^{*}$ &
                \cellcolor{magenta!5}\textbf{75.3} & \cellcolor{magenta!5}\textbf{91.4} &
                \cellcolor{magenta!5}55.3$^{*}$ & \cellcolor{magenta!5}65.5$^{*}$ & \cellcolor{magenta!5}79.8$^{*}$ &
                \cellcolor{magenta!5}{75.2}$^{\dagger}$ & \cellcolor{magenta!5}{91.4}$^{\dagger}$ \\

        \\[-1ex] & \textit{PaliGemma-3B-pt-224-ft}~$^{\oplus}$ & 3B &
            $Greedy$ &
                42.8 & 54.1 & 74.1 &
                39.7 & 50.7 & 70.1 &
                -- & -- &
                \textbf{48.1} & \textbf{60.1} & \textbf{76.6} &
                -- & -- \\

            &~\cite{beyer2024paligemmaversatile3bvlm} &&
            $RepeatRot$ & 
                 42.8 &  54.3 &  74.0 &
                 40.0 &  51.1 &  70.3 &
                 -- &  -- &
                 47.9 &  59.9&  76.3& 
                 -- &  --\\
                
            &&&$RandomRot$ & 
                 45.7 &  57.3  &  73.8 &
                 41.7 &  53.4 &  70.4 &
                 -- &  -- &
                 46.0 &  57.2 &  73.4 & 
                 -- &  -- \\
                
            &&&\cellcolor{magenta!5}\textcolor{magenta}{\textit{\textbf{ROSA}}} & 
                \cellcolor{magenta!5}\textbf{53.3} & \cellcolor{magenta!5}\textbf{65.5} & \cellcolor{magenta!5}\textbf{79.3} &
                \cellcolor{magenta!5}\textbf{47.5} & \cellcolor{magenta!5}\textbf{59.5} & \cellcolor{magenta!5}\textbf{75.6} &
                \cellcolor{magenta!5}-- & \cellcolor{magenta!5}-- &
                \cellcolor{magenta!5}47.6 & \cellcolor{magenta!5}59.5 & \cellcolor{magenta!5}75.5 &
                \cellcolor{magenta!5}-- & \cellcolor{magenta!5}--\\

        \\[-1ex] & \textit{PaliGemma-3B-224-ft-ocrvqa} & 3B &
            $Greedy$ &
                -- & -- &-- &
                -- & -- & -- &
                63.2 & 84.1 &
                -- & -- & -- &
                74.1 & 90.5 \\

            &~\cite{beyer2024paligemmaversatile3bvlm}&&
            $RepeatRot$ & 
                 -- &  --&  -- &
                 -- &  --&  -- &
                 63.3 &  84.3 &
                 -- &  -- &  -- & 
                 \textbf{74.3} &  \textbf{90.9} \\
                
            &&&$RandomRot$ & 
                 -- & -- &  -- &
                 -- & -- &  -- &
                 66.6 &  85.8 &
                 -- &  -- &  -- &
                 72.1 &  88.7 \\
                
            &&&\cellcolor{magenta!5}\textcolor{magenta}{\textit{\textbf{ROSA}}} & 
                \cellcolor{magenta!5}-- & \cellcolor{magenta!5}-- & \cellcolor{magenta!5}-- &
                \cellcolor{magenta!5}-- & \cellcolor{magenta!5}-- & \cellcolor{magenta!5}-- &
                \cellcolor{magenta!5}\textbf{71.3} & \cellcolor{magenta!5}\textbf{88.3} &
                \cellcolor{magenta!5}-- & \cellcolor{magenta!5}-- & \cellcolor{magenta!5}-- &
                \cellcolor{magenta!5}{71.3}$^{\dagger}$ & \cellcolor{magenta!5}{88.6}$^{\dagger}$ \\
                
    \hline & \\[-1ex]
                
    \multirow{4}{*}{\modelcategory{API call\\only}} & 
        \textit{GPT-4o-mini} & N/A &
            $Greedy$ &
                33.1$^{*}$ & 46.0$^{*}$ & 71.9$^{*}$ &  
                32.8 & 43.5 & 69.4 & 
                45.7 & 76.5 &  
                37.1$^{*}$ & 48.4$^{*}$ & \textbf{74.2}$^{*}$ & 
                50.4 & \textbf{79.9} \\
            
            &~\cite{openai2024gpt4technicalreport}&&
            $RepeatRot$ & 
                 39.1$^{*}$ &  52.3$^{*}$ &  76.9$^{*}$ &
                 33.4 &  44.7 &  69.6 &
                 46.6 &  76.9 &
                 38.3$^{*}$ &  49.6$^{*}$  &  74.4$^{*}$ & 
                 50.2 &  79.6 \\
            
            &&&$RandomRot$ & 
                 38.0$^{*}$ &  51.1$^{*}$ &  75.7$^{*}$ &
                 36.7$^{*}$ &  48.4$^{*}$ &  71.4$^{*}$ &
                 49.3 &  78.3&
                 \textbf{38.9}$^{*}$ &  50.5$^{*}$ & 73.3 $^{*}$ &
                 \textbf{50.6} &  79.1 \\

            &&&\cellcolor{magenta!5}\textcolor{magenta}{\textit{\textbf{ROSA}}} & 
                \cellcolor{magenta!5}\textbf{41.2}$^{*}$ & \cellcolor{magenta!5}\textbf{55.3}$^{*}$ & \cellcolor{magenta!5}\textbf{77.8}$^{*}$ &
                \cellcolor{magenta!5}\textbf{38.7}$^{*}$ & \cellcolor{magenta!5}\textbf{51.5}$^{*}$ & \cellcolor{magenta!5}\textbf{73.8}$^{*}$ &
                \cellcolor{magenta!5}\textbf{50.1} & \cellcolor{magenta!5}\textbf{79.0} &
                \cellcolor{magenta!5}38.4$^{*}$ & \cellcolor{magenta!5}\textbf{50.8}$^{*}$ & \cellcolor{magenta!5}73.6$^{*}$ &
                \cellcolor{magenta!5}49.8 & \cellcolor{magenta!5}79.2 \\
    
    \hline
    \multicolumn{4}{c}{} &
    &&&
    &&&
    &&
    &&&
    & \\[-1ex]
    \multicolumn{4}{c}{Median Absolute $\Delta$ ($ROSA - Greedy$)} &
    \contour{black}{\scriptsize$\uparrow$}~10.3 & 
    \contour{black}{\scriptsize$\uparrow$}~11.4 & 
    \contour{black}{\scriptsize$\uparrow$}~5.9 
    &
    \contour{black}{\scriptsize$\uparrow$}~7.8 &
    \contour{black}{\scriptsize$\uparrow$}~8.8 &
    \contour{black}{\scriptsize$\uparrow$}~5.3 
    &
    \contour{black}{\scriptsize$\uparrow$}~7.4 & 
    \contour{black}{\scriptsize$\uparrow$}~4.2 
    &
    $\downarrow$~(-0.5) &
    $\downarrow$~(-0.6) & 
    $\downarrow$~(-1.1) 
    &
    $\downarrow$~(-0.2) &
    $\downarrow$~(-0.6) \\
  
\end{NiceTabular}
    }
    \endgroup
    \caption{Performance comparison of several state-of-the-art multimodal models using the $Greedy$ approach and our $ROSA$ strategy, along with two $ROSA$-derived baselines: $RepeatRot$ (no rotations, same number of inferences) and $RandomRot$ (random rotations including the original orientation). 
    Models were evaluated on \textit{Exact Match} (EM), \textit{Semantic Answer Similarity} (SAS), and \textit{Standard VQA Accuracy} (Acc), grouped by datasets with correctly and incorrectly oriented text. 
    $Acc$ is not computed for \textit{OCR-VQA} datasets due to their single reference answer. ($\oplus$) Pretrained model was fine-tuned on \textit{VizWiz-VQA}, excluding \textit{VizWiz-Text} samples.
    \textbf{Bold} highlights the best result per dataset/strategy. 
    ($*$) Results may be influenced by model exposure to the dataset during pretraining/fine-tuning. 
    ($\dagger$) Results are averaged over five runs, with maximum recorded standard deviations of \textit{0.2} for EM and \textit{0.2} for SAS, as these values are meant for comparison with oracle settings.
    ($-$) Denotes non-applicable values for fine-tuned models. ($N/A$) Indicate no exact parameter count provided. 
    \textit{Results show that $ROSA$ consistently outperforms $Greedy$ in misoriented text scenarios while achieving competitive performance under oracle conditions.}}
    \label{tab:strategy_performance}
    \vspace{-1em}
\end{table*}

Table~\ref{tab:strategy_performance} compares the performance of several state-of-the-art multimodal models using the $Greedy$ decoding approach and our proposed $ROSA$ strategy. Models were evaluated on multiple datasets using \textit{Exact Match} (EM), \textit{Standard VQA Accuracy} (Acc), and \textit{Semantic Answer Similarity} (SAS). The results are structured into two categories: datasets containing images with incorrectly oriented---rotated---text and those with correctly oriented text.

The first clear pattern we observe is that all models struggle significantly when faced with incorrectly oriented text datasets.
Specifically, \textit{VizWiz\,$(Random)$} and the reference dataset \textit{OCR-VQA\,$(Random)$} show an average EM performance drop of approximately \textbf{7.7 points} compared to their correctly oriented counterparts. These results, obtained using the \textit{Greedy} approach, highlight a fundamental weakness of current models in handling non-standard orientations, reinforcing our initial hypothesis.

Across all datasets with incorrectly rotated text, $ROSA$ consistently outperforms $Greedy$ strategy, demonstrating its robustness in handling challenging text orientations. Notably, in \textit{VizWiz\,$(Conventional)$}---a dataset that reflect capture patterns derived from visually impaired people experiences---$ROSA$ achieves an improvement of up to \textbf{11.7 points} in EM.
Additionally, in datasets with artificially rotated text, such as \textit{VizWiz\,$(Random)$} and \textit{OCR-VQA\,$(Random)$}, $ROSA$ also proved effective, leading to EM increases of up to \textbf{11 points} and \textbf{17 points}, respectively.

To analyze the specific factors behind $ROSA$'s improvements, we introduced two controlled $ROSA$-derived baselines. The first, \textit{RepeatRot}, isolates the effect of multiple sampling by preserving the original, unrotated image orientation while maintaining the same generation settings as the full strategy. The second, \textit{RandomRot}, assesses the impact of applying random rotations between 0º and 360º instead of the structured 90º, 180º, and 270º variants. 

When comparing $ROSA$ to \textit{RepeatRot}, we find that the latter consistently underperforms, confirming that the observed improvements are primarily driven by the visual variability introduced through canonical rotations, rather than by sampling alone. Similarly, when compared to \textit{RandomRot}, we find that $ROSA$'s structured approach yields superior results, suggesting that fixed 90º, 180º, and 270º rotations provide a more effective coverage of the image space than arbitrary transformations.

For datasets with correctly oriented text, a different trend emerges: the $Greedy$ strategy generally outperforms $ROSA$. To investigate this drop, we examined the \textit{RepeatRot} baseline, which ---similar to its behavior in incorrectly rotated text datasets--- performs better than $Greedy$ in this scenario.
This leads to two key insights. First, sampling alone offers advantages over conventional $Greedy$ decoding, likely by introducing controlled variability that enhances answer diversity. Second, one reason for $ROSA$'s lower performance on these datasets could be the introduction of unnecessary rotations, which distort the text orientation and add variability, negatively affecting confidence in the correct answers. In contrast, $Greedy$ processes the text optimally, without these alterations.
Regarding \textit{RandomRot}, no clear pattern is evident in relation to the $ROSA$ strategy, suggesting that in scenarios with correctly oriented text sets, neither approach provides a clear advantage.

To establish an upper bound on model performances and assess $ROSA$'s effectiveness, we define oracle values as the results obtained by applying the $Greedy$ approach to the \textit{OCR-VQA\,$(Oriented)$} dataset. Unlike \textit{VizWiz-Text\,($Oriented$)}, which may include text with slight inclinations due to the criteria used to define the thresholds for correctly oriented text, \textit{OCR-VQA\,$(Oriented)$} serves as an ideal reference point, as it consists entirely of images with perfectly aligned text. 
This dataset simulates the outcome of a hypothetical pipeline with a perfect text detector that accurately locates and aligns text before inference, ensuring optimal conditions for model performance. In this idealized scenario, text orientation is no longer a limiting factor, making the $Greedy$ strategy's performance on \textit{OCR-VQA\,$(Oriented)$} a reliable benchmark for the highest achievable accuracy under optimal preprocessing conditions.

We evaluated the $ROSA$ on the three models that achieved the highest performance under the oracle configurations. To ensure the robustness and consistency of the results, $ROSA$'s performance in these settings was averaged over five independent runs. 
As indicated by ``$\dagger$'' in Table~\ref{tab:strategy_performance}, $ROSA$'s results closely approach oracle EM performance, even surpassing it in one case.  However, it is important to note that these values do not account for the inherent errors of a text detector in real-world scenarios, which would likely reduce the oracle’s performance and further increase the advantage of $ROSA$ in practical applications.

\subsection{Detailed performance analysis of ROSA}
To better understand the impact of our strategy and the models' ability to handle text at different angles, we analyzed the contribution of correct answers from each rotation used by $ROSA$. This analysis was conducted on the \textit{VizWiz\,$(Conventional)$} dataset, which is particularly relevant to our study as it contains challenging images with text rotations that reflect the capture patterns of people with visual impairments. 

For each model, we calculated the percentage of final correct predictions from $ROSA$ attributable to each rotation used in its response selection process, using EM as the evaluation criterion. 

\begin{figure}[!t]
    \centering
    \includegraphics[width=\linewidth]{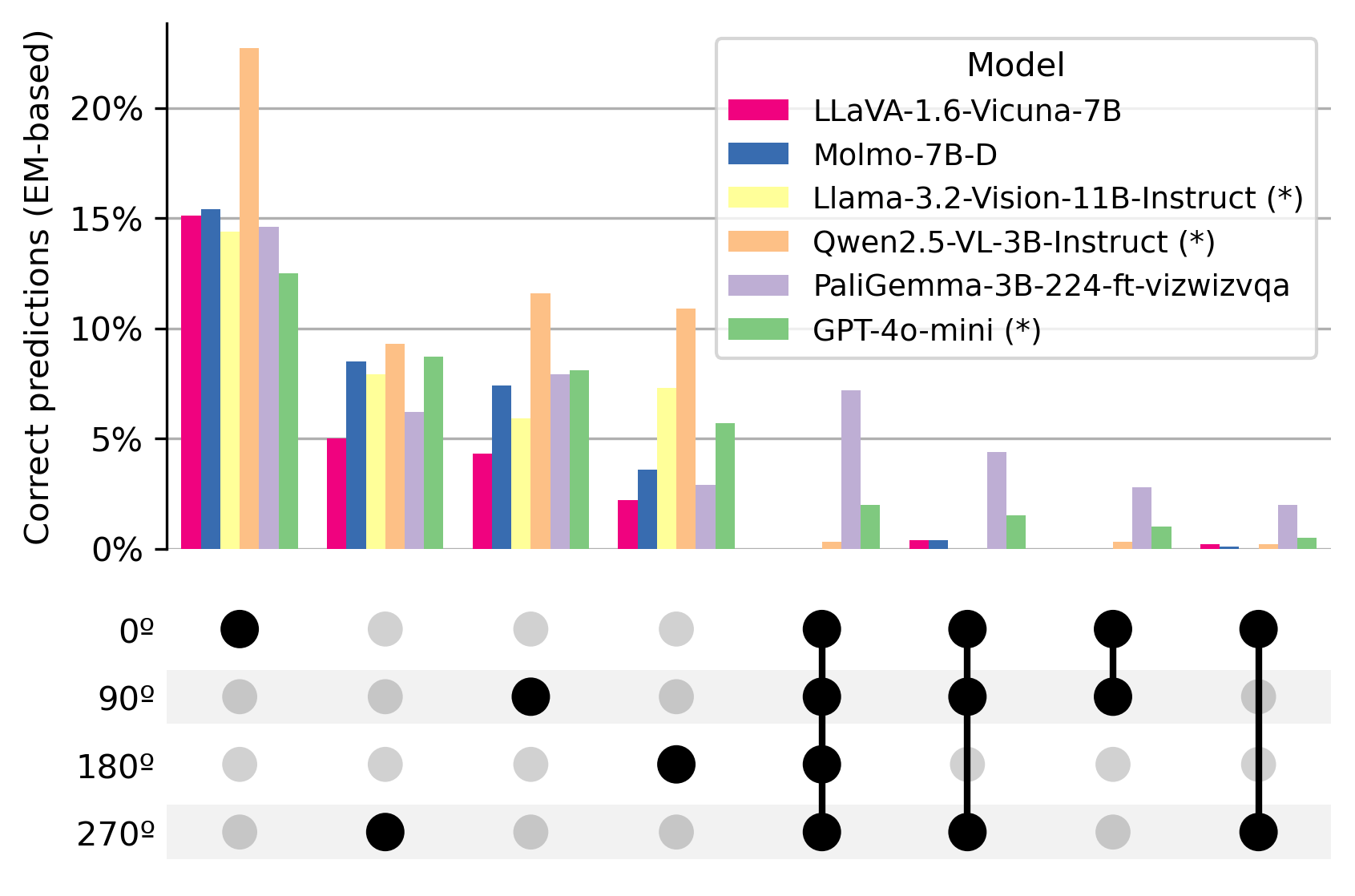}
    \caption{Distribution of final correct EM-based predictions from $ROSA$ attributable to each rotation used, for the \textit{VizWiz\,$(Conventional)$} dataset. Rotation 0º is defined as the image orientation with correctly oriented text. The upper bars show the contribution size per model, and the filled points indicate which rotations contributed to the correct final answer. $(*)$ Models may have pre/fine-tuning data exposure.
    \textit{Perhaps unsurprisingly, models perform better on images with correctly oriented text (0º) and less accurately on upside-down text (180º).}}
    \label{fig:rosa_correctnes_upsetplot}
    \vspace{-1em}
\end{figure}

The results are presented in the UpSet plot in Figure~\ref{fig:rosa_correctnes_upsetplot}, where the contribution of each rotation used by $ROSA$ is shown as a matrix of intersections. To facilitate the understanding of the plot, rotation 0º was defined as the orientation of the image where the text is correctly oriented, and the other rotations are applied relative to this. Each row at the bottom corresponds to an applied rotation, while each column represents a non-empty set of rotation combinations. The upper bars indicate the percentage size of these contributions, grouped by each evaluated model, and the filled points indicate which specific rotations contributed to generating a final correct answer by $ROSA$. Combinations of rotations where the sum of contributions to each model is below 3\% were excluded from the plot.

The first column shows that most of $ROSA$'s correct predictions come from images with correctly oriented text. This observation not only supports the logic behind the design of our strategy but also highlights the inherent preference of the models for this optimal orientation condition. On the other hand, the fourth column (180º) shows the loss of precision in images with upside-down text.
\section{Conclusions}
\label{sec:conclusions}

Throughout this work, we have demonstrated that answering visual questions requiring the reading and detection of incorrectly oriented text in images poses a significant challenge for the current multimodal models. In particular, we studied its impact on images captured by individuals with visual impairments during their daily activities. Through interviews with members of this population, we identified unique patterns and learned conventions that influence how images are taken, which allowed us to develop tailored evaluation datasets that reflect these challenges.

Our proposed strategy, $ROSA$, effectively mitigated the impact of text rotation, consistently improving $Greedy$ decoding performance across all tested models. We therefore recommend its adoption in datasets where a high proportion of images are known to contain such characteristics.

These findings underscore the importance of accounting for the specific needs of people with visual impairments when designing and evaluating future VQA models, particularly in assistive contexts where accuracy and reliability are critical.
\section*{Limitations}
\label{sec:limitations}

We evaluated our strategy on images with adequate quality and clearly identifiable content to answer the associated questions. However, many real-world images captured by visually impaired individuals exhibit blurriness, which hinders information extraction. As our strategy does not explicitly address such conditions, its applicability in uncontrolled scenarios may be limited. Before deploying it in assistive systems, it is crucial to assess the uncertainty of model predictions. Methods such as those proposed by \citet{eisenschlos-etal-2024-selectively} could allow models to estimate confidence in advance and, when necessary, abstain or request clarification, thereby improving their reliability in critical applications.

The effectiveness of our approach is also influenced by the models employed, as it cannot fully overcome intrinsic limitations in interpreting certain questions or reading text in images. Performance further depends on model size, which in our experiments was constrained by available computational resources. Moreover, $ROSA$ requires multiple inferences per image, leading to increased processing time compared to standard inference methods. While this additional cost may be justified by performance gains, it can be a limitation in real-time applications or resource-constrained environments.
\section*{Ethical considerations}
\label{sec:ethics}

This work involved interviews with visually impaired individuals, all conducted with informed consent. We also used the \textit{VizWiz-VQA} dataset, whose creators ensured appropriate consent procedures and anonymization of personal information.

Given the sensitivity of working with this population, we recognize the ethical implications and potential risks of applying our findings in real-world assistive technologies ---particularly because end users may face difficulties in verifying the correctness of the outputs. For this reason, a thorough assessment of potential failure modes is essential before any practical deployment.
\section*{Acknowledgments}

We would like to express our sincere gratitude to Vanesa Fundaró, Maximiliano González, Mariano Javier López, Lucas Emiliano Martin, Horacio Maximiliano Noble, Marco Lione Stuto, and Mercedes Sara Yanes, who generously participated in this work. Their testimonies and life experiences were essential for a deep understanding of the problem addressed.

We are especially grateful to Mercedes Hug and Fernando Bermejo for their valuable interdisciplinary contributions, which were key to the design of the interviews and the development of the related activities.

We also extend our thanks to the institutions that played a key role in helping us establish a connection with the interview participants, including the \textit{Fundación GAUDE}~\footnote{https://www.facebook.com/FUNDACIONGAUDE}---a private center for rehabilitation and educational integration for people with visual impairments--- and the \textit{Instituto Helen Keller}~\footnote{https://www.facebook.com/institutohelenkellercba}---a special education school under the Ministerio de Educación de la Provincia de Córdoba--- both located in the city of Córdoba, Argentina.

This work used computational resources from UNC Supercómputo (CCAD) – Universidad Nacional de Córdoba~\footnote{https://supercomputo.unc.edu.ar}, which are part of SNCAD, República Argentina. It was also supported by the computing power of Nodo de Cómputo IA, from Ministerio de Ciencia y Tecnología de la Provincia de Córdoba, located in San Francisco, Argentina. We especially thank Nicolás Wolovick for his unwavering and unconditional support.

Finally, we would like to extend our appreciation to all those who, directly or indirectly, contributed to the development and completion of this work.

\bibliography{bib/custom}

\begin{thebibliography}{35}
\expandafter\ifx\csname natexlab\endcsname\relax\def\natexlab#1{#1}\fi

\bibitem[{Antol et~al.(2015)Antol, Agrawal, Lu, Mitchell, Batra, Zitnick, and Parikh}]{Antol_2015_ICCV}
Stanislaw Antol, Aishwarya Agrawal, Jiasen Lu, Margaret Mitchell, Dhruv Batra, C.~Lawrence Zitnick, and Devi Parikh. 2015.
\newblock \href {https://openaccess.thecvf.com/content_iccv_2015/html/Antol_VQA_Visual_Question_ICCV_2015_paper.html} {{VQA: Visual Question Answering}}.
\newblock In \emph{Proceedings of the IEEE International Conference on Computer Vision (ICCV)}.

\bibitem[{Bazazian et~al.(2017)Bazazian, Gomez, Nicolaou, Gomez, Karatzas, and Bagdanov}]{bazazian2017improvingtextproposalsscene}
Dena Bazazian, Raul Gomez, Anguelos Nicolaou, Lluis Gomez, Dimosthenis Karatzas, and Andrew~D. Bagdanov. 2017.
\newblock \href {http://arxiv.org/abs/1702.05089} {{Improving Text Proposals for Scene Images with Fully Convolutional Networks}}.

\bibitem[{Beyer et~al.(2024)Beyer, Steiner, Pinto, Kolesnikov, Wang, Salz, Neumann, Alabdulmohsin, Tschannen, Bugliarello, Unterthiner, Keysers, Koppula, Liu, Grycner, Gritsenko, Houlsby, Kumar, Rong, Eisenschlos, Kabra, Bauer, Bošnjak, Chen, Minderer, Voigtlaender, Bica, Balazevic, Puigcerver, Papalampidi, Henaff, Xiong, Soricut, Harmsen, and Zhai}]{beyer2024paligemmaversatile3bvlm}
Lucas Beyer, Andreas Steiner, André~Susano Pinto, Alexander Kolesnikov, Xiao Wang, Daniel Salz, Maxim Neumann, Ibrahim Alabdulmohsin, Michael Tschannen, Emanuele Bugliarello, Thomas Unterthiner, Daniel Keysers, Skanda Koppula, Fangyu Liu, Adam Grycner, Alexey Gritsenko, Neil Houlsby, Manoj Kumar, Keran Rong, Julian Eisenschlos, Rishabh Kabra, Matthias Bauer, Matko Bošnjak, Xi~Chen, Matthias Minderer, Paul Voigtlaender, Ioana Bica, Ivana Balazevic, Joan Puigcerver, Pinelopi Papalampidi, Olivier Henaff, Xi~Xiong, Radu Soricut, Jeremiah Harmsen, and Xiaohua Zhai. 2024.
\newblock \href {http://arxiv.org/abs/2407.07726} {{PaliGemma: A versatile 3B VLM for transfer}}.

\bibitem[{Bigham et~al.(2010)Bigham, Jayant, Ji, Little, Miller, Miller, Miller, Tatarowicz, White, White, and Yeh}]{VizWiz:nearly-real-time-answers}
Jeffrey~P. Bigham, Chandrika Jayant, Hanjie Ji, Greg Little, Andrew Miller, Robert~C. Miller, Robin Miller, Aubrey Tatarowicz, Brandyn White, Samual White, and Tom Yeh. 2010.
\newblock \href {https://doi.org/10.1145/1866029.1866080} {{V}iz{W}iz: nearly real-time answers to visual questions}.
\newblock In \emph{Proceedings of the 23nd Annual ACM Symposium on User Interface Software and Technology}, UIST '10, page 333–342, New York, NY, USA. Association for Computing Machinery.

\bibitem[{Biten et~al.(2019)Biten, Tito, Mafla, Gomez, Rusiñol, Jawahar, Valveny, and Karatzas}]{ST-VQA:dataset}
Ali~Furkan Biten, Rubèn Tito, Andrés Mafla, Lluis Gomez, Marçal Rusiñol, C.V. Jawahar, Ernest Valveny, and Dimosthenis Karatzas. 2019.
\newblock \href {https://doi.org/10.1109/ICCV.2019.00439} {{Scene Text Visual Question Answering}}.
\newblock In \emph{2019 IEEE/CVF International Conference on Computer Vision (ICCV)}, pages 4290--4300.

\bibitem[{Deitke et~al.(2024)Deitke, Clark, Lee, Tripathi, Yang, Park, Salehi, Muennighoff, Lo, Soldaini, Lu, Anderson, Bransom, Ehsani, Ngo, Chen, Patel, Yatskar, Callison-Burch, Head, Hendrix, Bastani, VanderBilt, Lambert, Chou, Chheda, Sparks, Skjonsberg, Schmitz, Sarnat, Bischoff, Walsh, Newell, Wolters, Gupta, Zeng, Borchardt, Groeneveld, Nam, Lebrecht, Wittlif, Schoenick, Michel, Krishna, Weihs, Smith, Hajishirzi, Girshick, Farhadi, and Kembhavi}]{deitke2024molmopixmoopenweights}
Matt Deitke, Christopher Clark, Sangho Lee, Rohun Tripathi, Yue Yang, Jae~Sung Park, Mohammadreza Salehi, Niklas Muennighoff, Kyle Lo, Luca Soldaini, Jiasen Lu, Taira Anderson, Erin Bransom, Kiana Ehsani, Huong Ngo, YenSung Chen, Ajay Patel, Mark Yatskar, Chris Callison-Burch, Andrew Head, Rose Hendrix, Favyen Bastani, Eli VanderBilt, Nathan Lambert, Yvonne Chou, Arnavi Chheda, Jenna Sparks, Sam Skjonsberg, Michael Schmitz, Aaron Sarnat, Byron Bischoff, Pete Walsh, Chris Newell, Piper Wolters, Tanmay Gupta, Kuo-Hao Zeng, Jon Borchardt, Dirk Groeneveld, Crystal Nam, Sophie Lebrecht, Caitlin Wittlif, Carissa Schoenick, Oscar Michel, Ranjay Krishna, Luca Weihs, Noah~A. Smith, Hannaneh Hajishirzi, Ross Girshick, Ali Farhadi, and Aniruddha Kembhavi. 2024.
\newblock \href {http://arxiv.org/abs/2409.17146} {{Molmo and PixMo: Open Weights and Open Data for State-of-the-Art Vision-Language Models}}.

\bibitem[{Eisenschlos et~al.(2024)Eisenschlos, Maina, Ivetta, and Benotti}]{eisenschlos-etal-2024-selectively}
Julian Eisenschlos, Hern{\'a}n Maina, Guido Ivetta, and Luciana Benotti. 2024.
\newblock \href {https://doi.org/10.18653/v1/2024.findings-acl.250} {Selectively answering visual questions}.
\newblock In \emph{Findings of the Association for Computational Linguistics: ACL 2024}, pages 4219--4229, Bangkok, Thailand. Association for Computational Linguistics.

\bibitem[{Engstrom et~al.(2019)Engstrom, Tran, Tsipras, Schmidt, and Madry}]{engstrom2019a}
Logan Engstrom, Brandon Tran, Dimitris Tsipras, Ludwig Schmidt, and Aleksander Madry. 2019.
\newblock \href {https://openreview.net/forum?id=BJfvknCqFQ} {A rotation and a translation suffice: Fooling {CNN}s with simple transformations}.

\bibitem[{Gal and Ghahramani(2016)}]{pmlr-v48-gal16}
Yarin Gal and Zoubin Ghahramani. 2016.
\newblock \href {https://proceedings.mlr.press/v48/gal16.html} {Dropout as a bayesian approximation: Representing model uncertainty in deep learning}.
\newblock In \emph{Proceedings of The 33rd International Conference on Machine Learning}, volume~48 of \emph{Proceedings of Machine Learning Research}, pages 1050--1059, New York, New York, USA. PMLR.

\bibitem[{Gurari et~al.(2018)Gurari, Li, Stangl, Guo, Lin, Grauman, Luo, and Bigham}]{gurari2018vizwiz}
Danna Gurari, Qing Li, Abigale~J. Stangl, Anhong Guo, Chi Lin, Kristen Grauman, Jiebo Luo, and Jeffrey~P. Bigham. 2018.
\newblock \href {https://doi.org/10.1109/CVPR.2018.00380} {{V}iz{W}iz {G}rand {C}hallenge: {A}nswering {V}isual {Q}uestions from {B}lind {P}eople}.
\newblock In \emph{2018 IEEE/CVF Conference on Computer Vision and Pattern Recognition}, pages 3608--3617.

\bibitem[{He et~al.(2016)He, Huang, Qiao, and Yao}]{he2016accuratetextlocalizationnatural}
Tong He, Weilin Huang, Yu~Qiao, and Jian Yao. 2016.
\newblock \href {http://arxiv.org/abs/1603.09423} {{Accurate Text Localization in Natural Image with Cascaded Convolutional Text Network}}.

\bibitem[{He et~al.(2018)He, Zhang, Yin, and Liu}]{he2018MOSceneText}
Wenhao He, Xu-Yao Zhang, Fei Yin, and Cheng-Lin Liu. 2018.
\newblock \href {https://doi.org/10.1109/TIP.2018.2855399} {{Multi-Oriented and Multi-Lingual Scene Text Detection With Direct Regression}}.
\newblock \emph{IEEE Transactions on Image Processing}, 27(11):5406--5419.

\bibitem[{Hu et~al.(2020)Hu, Singh, Darrell, and Rohrbach}]{hu2020iterative}
Ronghang Hu, Amanpreet Singh, Trevor Darrell, and Marcus Rohrbach. 2020.
\newblock \href {https://doi.org/10.1109/CVPR42600.2020.01001} {{ Iterative Answer Prediction With Pointer-Augmented Multimodal Transformers for TextVQA }}.
\newblock In \emph{2020 IEEE/CVF Conference on Computer Vision and Pattern Recognition (CVPR)}, pages 9989--9999, Los Alamitos, CA, USA. IEEE Computer Society.

\bibitem[{Krizhevsky et~al.(2012)Krizhevsky, Sutskever, and Hinton}]{krizhevsky-et-al-2012-imagenet}
Alex Krizhevsky, Ilya Sutskever, and Geoffrey~E Hinton. 2012.
\newblock \href {https://proceedings.neurips.cc/paper_files/paper/2012/file/c399862d3b9d6b76c8436e924a68c45b-Paper.pdf} {Imagenet classification with deep convolutional neural networks}.
\newblock In \emph{Advances in Neural Information Processing Systems}, volume~25. Curran Associates, Inc.

\bibitem[{Lee et~al.(2023)Lee, Joshi, Turc, Hu, Liu, Eisenschlos, Khandelwal, Shaw, Chang, and Toutanova}]{lee2023pix2struct}
Kenton Lee, Mandar Joshi, Iulia Turc, Hexiang Hu, Fangyu Liu, Julian Eisenschlos, Urvashi Khandelwal, Peter Shaw, Ming-Wei Chang, and Kristina Toutanova. 2023.
\newblock \href {http://arxiv.org/abs/2210.03347} {Pix2struct: Screenshot parsing as pretraining for visual language understanding}.

\bibitem[{Lewis(1969)}]{Lewis1969-LEWCAP-4}
David~Kellogg Lewis. 1969.
\newblock \emph{Convention: A Philosophical Study}.
\newblock Wiley-Blackwell, Cambridge, MA, USA.

\bibitem[{Liao et~al.(2018)Liao, Zhu, Shi, Xia, and Bai}]{liao2018RSROSceneTextR}
Minghui Liao, Zhen Zhu, Baoguang Shi, Gui-song Xia, and Xiang Bai. 2018.
\newblock \href {https://doi.org/10.1109/CVPR.2018.00619} {Rotation-sensitive regression for oriented scene text detection}.
\newblock In \emph{2018 IEEE/CVF Conference on Computer Vision and Pattern Recognition}, pages 5909--5918.

\bibitem[{Liu et~al.(2024{\natexlab{a}})Liu, Li, Li, and Lee}]{Liu_2024_CVPR}
Haotian Liu, Chunyuan Li, Yuheng Li, and Yong~Jae Lee. 2024{\natexlab{a}}.
\newblock \href {https://doi.org/10.1109/CVPR52733.2024.02484} {Improved baselines with visual instruction tuning}.
\newblock In \emph{2024 IEEE/CVF Conference on Computer Vision and Pattern Recognition (CVPR)}, pages 26296--26306.

\bibitem[{Liu et~al.(2024{\natexlab{b}})Liu, Li, Li, Li, Zhang, Shen, and Lee}]{liu2024llavanext}
Haotian Liu, Chunyuan Li, Yuheng Li, Bo~Li, Yuanhan Zhang, Sheng Shen, and Yong~Jae Lee. 2024{\natexlab{b}}.
\newblock \href {https://llava-vl.github.io/blog/2024-01-30-llava-next/} {{LLaVA-NeXT: Improved reasoning, OCR, and world knowledge}}.

\bibitem[{Ludíková and Finková(2012)}]{LUDIKOVA2012971}
Libuše Ludíková and Dita Finková. 2012.
\newblock \href {https://doi.org/https://doi.org/10.1016/j.sbspro.2012.09.587} {{Improvement in Education of People with Visual Impairment}}.
\newblock \emph{Procedia - Social and Behavioral Sciences}, 55:971--979.
\newblock 3rd. International Conference on New Horizons in Education - INTE 2012.

\bibitem[{Ma et~al.(2018)Ma, Shao, Ye, Wang, Wang, Zheng, and Xue}]{Ma2018AOSceneTextD}
Jianqi Ma, Weiyuan Shao, Hao Ye, Li~Wang, Hong Wang, Yingbin Zheng, and Xiangyang Xue. 2018.
\newblock \href {https://doi.org/10.1109/TMM.2018.2818020} {{Arbitrary-Oriented Scene Text Detection via Rotation Proposals}}.
\newblock \emph{IEEE Transactions on Multimedia}, 20(11):3111--3122.

\bibitem[{Majerova(2017)}]{MAJEROVA2017751}
Hana Majerova. 2017.
\newblock \href {https://doi.org/https://doi.org/10.1016/j.sbspro.2017.02.117} {{The Person in a Situation of Visual Impairment and its Perception and Imagination from the Qualitative Viewpoint}}.
\newblock \emph{Procedia - Social and Behavioral Sciences}, 237:751--757.
\newblock Education, Health and ICT for a Transcultural World.

\bibitem[{{Meta AI}(2024)}]{grattafiori2024llama3herdmodels}
{Meta AI}. 2024.
\newblock \href {http://arxiv.org/abs/2407.21783} {The llama 3 herd of models}.

\bibitem[{Mishra et~al.(2019)Mishra, Shekhar, Singh, and Chakraborty}]{mishraICDAR19}
Anand Mishra, Shashank Shekhar, Ajeet~Kumar Singh, and Anirban Chakraborty. 2019.
\newblock \href {https://doi.org/10.1109/ICDAR.2019.00156} {{OCR-VQA: Visual Question Answering by Reading Text in Images}}.
\newblock In \emph{2019 International Conference on Document Analysis and Recognition (ICDAR)}, pages 947--952.

\bibitem[{OpenAI(2024)}]{openai2024gpt4technicalreport}
OpenAI. 2024.
\newblock \href {http://arxiv.org/abs/2303.08774} {Gpt-4 technical report}.

\bibitem[{{Qwen Team}(2025)}]{qwen2.5-VL}
{Qwen Team}. 2025.
\newblock \href {https://qwenlm.github.io/blog/qwen2.5-vl/} {{Qwen2.5-VL}}.

\bibitem[{Shorten and Khoshgoftaar(2019)}]{shorten2019survey}
Connor Shorten and Taghi~M Khoshgoftaar. 2019.
\newblock \href {https://journalofbigdata.springeropen.com/articles/10.1186/s40537-019-0197-0#citeas} {A survey on image data augmentation for deep learning}.
\newblock \emph{Journal of Big Data}, 6(1):1--48.

\bibitem[{Simonyan and Zisserman(2015)}]{Simonyan15}
Karen Simonyan and Andrew Zisserman. 2015.
\newblock \href {https://www.robots.ox.ac.uk/~vgg/publications/2015/Simonyan15/} {Very deep convolutional networks for large-scale image recognition}.
\newblock In \emph{International Conference on Learning Representations}.

\bibitem[{Singh et~al.(2019)Singh, Natarajan, Shah, Jiang, Chen, Batra, Parikh, and Rohrbach}]{singh2019towards}
Amanpreet Singh, Vivek Natarajan, Meet Shah, Yu~Jiang, Xinlei Chen, Dhruv Batra, Devi Parikh, and Marcus Rohrbach. 2019.
\newblock \href {https://doi.org/10.1109/CVPR.2019.00851} {{Towards VQA Models That Can Read}}.
\newblock In \emph{2019 IEEE/CVF Conference on Computer Vision and Pattern Recognition (CVPR)}, pages 8309--8318.

\bibitem[{Wang et~al.(2020)Wang, Wei, Dong, Bao, Yang, and Zhou}]{wang2020minilm}
Wenhui Wang, Furu Wei, Li~Dong, Hangbo Bao, Nan Yang, and Ming Zhou. 2020.
\newblock \href {https://proceedings.neurips.cc/paper/2020/hash/3f5ee243547dee91fbd053c1c4a845aa-Abstract.html} {{MiniLM: Deep Self-Attention Distillation for Task-Agnostic Compression of Pre-Trained Transformers}}.
\newblock In \emph{Proceedings of the 34th International Conference on Neural Information Processing Systems}. Curran Associates Inc.

\bibitem[{Wang et~al.(2023)Wang, Wei, Schuurmans, Le, Chi, Narang, Chowdhery, and Zhou}]{wang2023selfconsistency}
Xuezhi Wang, Jason Wei, Dale Schuurmans, Quoc~V Le, Ed~H. Chi, Sharan Narang, Aakanksha Chowdhery, and Denny Zhou. 2023.
\newblock \href {https://openreview.net/forum?id=1PL1NIMMrw} {Self-consistency improves chain of thought reasoning in language models}.
\newblock In \emph{The Eleventh International Conference on Learning Representations}.

\bibitem[{Yang et~al.(2021)Yang, Lu, Wang, Yin, Florencio, Wang, Zhang, Zhang, and Luo}]{Yang2021TAP}
Zhengyuan Yang, Yijuan Lu, Jianfeng Wang, Xi~Yin, Dinei Florencio, Lijuan Wang, Cha Zhang, Lei Zhang, and Jiebo Luo. 2021.
\newblock Tap: Text-aware pre-training for text-vqa and text-caption.
\newblock In \emph{Proceedings of the IEEE/CVF Conference on Computer Vision and Pattern Recognition (CVPR)}, pages 8751--8761.

\bibitem[{Yao et~al.(2016)Yao, Bai, Sang, Zhou, Zhou, and Cao}]{yao2016SceneTextDH}
Cong Yao, Xiang Bai, Nong Sang, Xinyu Zhou, Shuchang Zhou, and Zhimin Cao. 2016.
\newblock \href {http://arxiv.org/abs/1606.09002} {{Scene Text Detection via Holistic, Multi-Channel Prediction}}.

\bibitem[{Zeng et~al.(2020)Zeng, Wang, Chiu, Bhattacharya, and Gurari}]{ZengWCBG20}
Xiaoyu Zeng, Yanan Wang, Tai-Yin Chiu, Nilavra Bhattacharya, and Danna Gurari. 2020.
\newblock \href {https://doi.org/10.1145/3415220} {{Vision Skills Needed to Answer Visual Questions}}.
\newblock \emph{Proceedings of the Association for Computing Machinery (ACM) on Human-Computer Interaction}, 4(CSCW2).

\bibitem[{Zhang et~al.(2024)Zhang, Yang, Zhu, and Yin}]{zhang2024ASTextD}
Shi-Xue Zhang, Chun Yang, Xiaobin Zhu, and Xu-Cheng Yin. 2024.
\newblock \href {https://doi.org/10.1109/TMM.2023.3286657} {{Arbitrary Shape Text Detection via Boundary Transformer}}.
\newblock \emph{IEEE Transactions on Multimedia}, 26:1747--1760.

\end{thebibliography}
\bibliographystyle{acl_natbib}
\end{document}